\documentclass[11pt,a4paper]{article}
\usepackage[hyperref]{acl2020}
\usepackage{times}
\usepackage{latexsym}

\usepackage{microtype}
\usepackage{algorithm}
\usepackage{algorithmic}
\usepackage{amsmath}
\usepackage{amsthm}
\usepackage{bbm}
\usepackage{graphicx}

\aclfinalcopy % Uncomment this line for the final submission
\title{Joint-training on Symbiosis Networks for Deep Nueral Machine Translation models}

\author{Zhengzhe Yu\textsuperscript{\rm 1}, 
  Jiaxin Guo\textsuperscript{\rm 1}, 
  Minghan Wang\textsuperscript{\rm 1}, 
  Daimeng Wei\textsuperscript{\rm 1},
  Hengchao Shang\textsuperscript{\rm 1},
  \\
  {\bf Zongyao Li\textsuperscript{\rm 1}},
  {\bf Zhanglin Wu\textsuperscript{\rm 1}},
  {\bf Yuxia Wang\textsuperscript{\rm 2}},
  {\bf Yimeng Chen\textsuperscript{\rm 1},}
  {\bf Chang Su\textsuperscript{\rm 1},}
  \\
  {\bf Min Zhang\textsuperscript{\rm 1},}
  {\bf Lizhi Lei\textsuperscript{\rm 1},}
  {\bf Shimin Tao\textsuperscript{\rm 1},}
  {\bf Hao Yang\textsuperscript{\rm 1}}
  \\
  \textsuperscript{\rm 1}Huawei Translation Services Center, Beijing, China \\
  \textsuperscript{\rm 2}The University of Melbourne, Melbourne, Australia\\
  \tt \{yuzhengzhe,guojiaxin1,wangminghan,weidaimeng,shanghengchao,\\ 
  \tt lizongyao,wuzhanglin2,chenyimeng,suchang8,zhangmin186,\\
  \tt leilizhi,taoshimin,yanghao30\}@huawei.com \\
  \texttt{yuxiaw@student.unimelb.edu.au}}

\date{}

\begin{document}
\maketitle
\begin{abstract}
Deep encoders have been proven to be effective in improving neural machine translation (NMT) systems, but it reaches the upper bound of translation quality when the number of encoder layers exceeds 18. Worse still, deeper networks consume a lot of memory, making it impossible to train efficiently. In this paper, we present Symbiosis Networks, which include a full network as the  Symbiosis Main Network (M-Net) and another shared sub-network with the same structure but less layers as the Symbiotic Sub Network (S-Net). We adopt Symbiosis Networks on Transformer-deep ($m$-$n$)\footnote{$m$ denotes the number of encoder sub-layers and $n$ denotes the number of decoder sub-layers in Transformer-deep ($m$-$n$)} architecture, and define a particular regularization loss $\mathcal{L}_{\tau}$ between the M-Net and S-Net in NMT. We apply joint-training on the Symbiosis Networks and aim to improve the M-Net performance. Our proposed training strategy improves Transformer-deep (12-6) by 0.61, 0.49 and 0.69 BLEU over the baselines under classic training on WMT'14 EN$\rightarrow$DE, DE$\rightarrow$EN and EN$\rightarrow$FR tasks. Furthermore, our Transformer-deep (12-6) even outperforms classic Transformer-deep (18-6).
\end{abstract}

\section{Introduction}
In recent years, neural models have led to state-of-the-art results in machine translation \citep{DBLP:conf/nips/SutskeverVL14,DBLP:journals/corr/BahdanauCB14,DBLP:conf/naacl/ShawUV18,DBLP:conf/naacl/YangMZWLZ21,DBLP:journals/corr/WuSCLNMKCGMKSJL16}. In particular, the Transformer model has become popular for its well-designed architecture and the ability to capture the dependency among positions over the entire sequence. There are two ways to improve the models performance, one is to use wider models and the other is to use deeper model.

\begin{figure}[htp]
\centering
\includegraphics[width=0.5\textwidth]{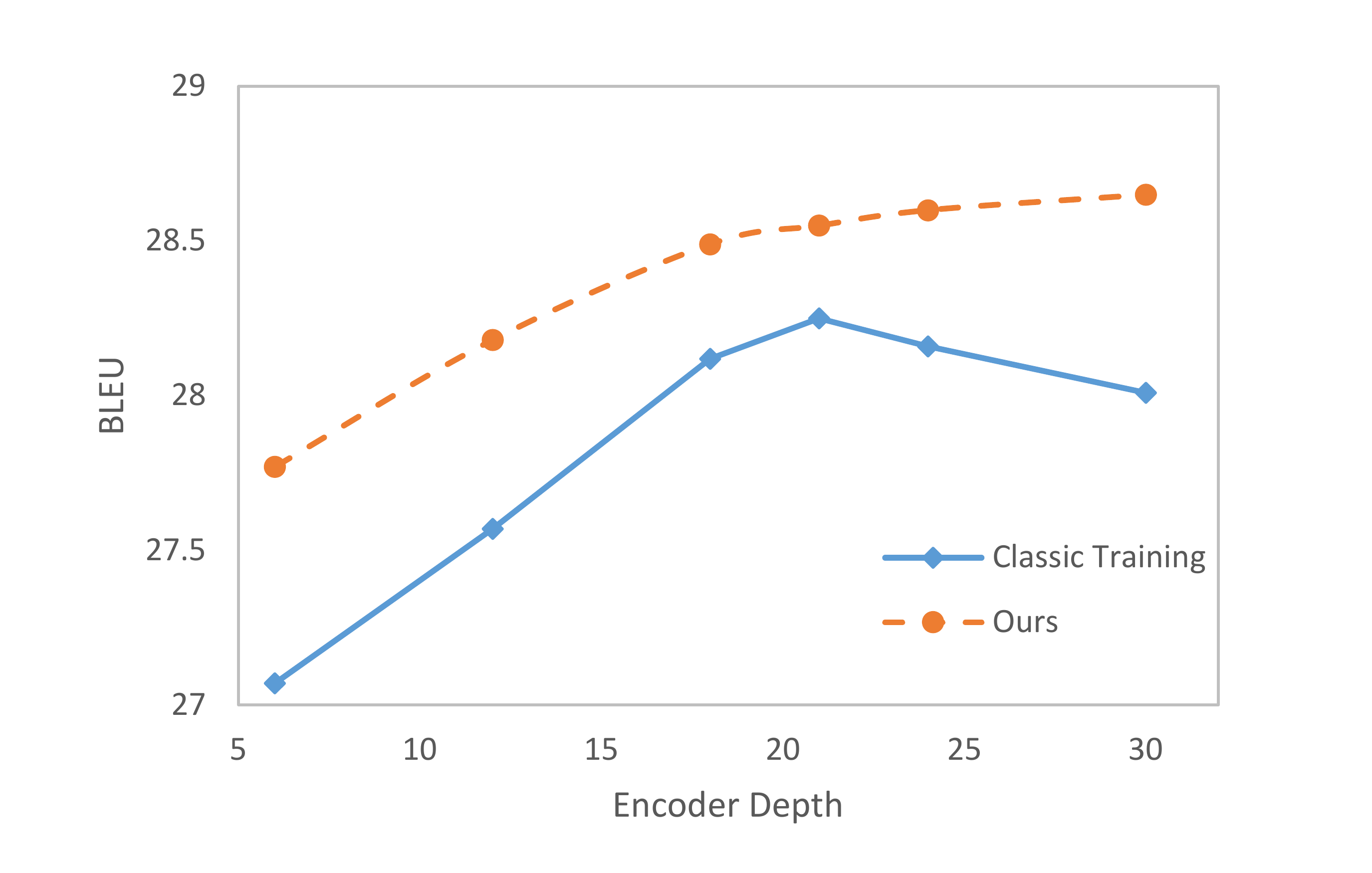} % Reduce the figure size so that it is slightly narrower than the column.
\caption{BLEU scores against the encoder depth under classic training and our proposed training on WMT'14 En$\leftrightarrow$De.}
\label{fig:bleu_with_depth}
\end{figure}

Deep model has shown promising BLEU improvements by either easing the information flow through the network or constraining the gradient norm across layers. But it reaches the upper bound of Translation quality when the number of encoder layers exceeds 18. See in Figure \ref{fig:bleu_with_depth}. Worse still, deeper networks consume a lot of memory, making it impossible to train efficiently. So, \textbf{how to improve the performance of deep model under limited layers is a meaningful question}.

In this paper, we study this question through training strategy. We present Symbiosis Networks on deep models, and apply joint-training on them. We define the margin gap between the Symbiosis Networks as a regularization loss. On WMT'14 EN$\rightarrow$DE, DE$\rightarrow$EN and EN$\rightarrow$FR tasks, our proposed training strategy can lead to significant improvements over the baselines widely.

\section{Background}

\begin{figure*}[htp]
\centering
\includegraphics[width=0.80\textwidth]{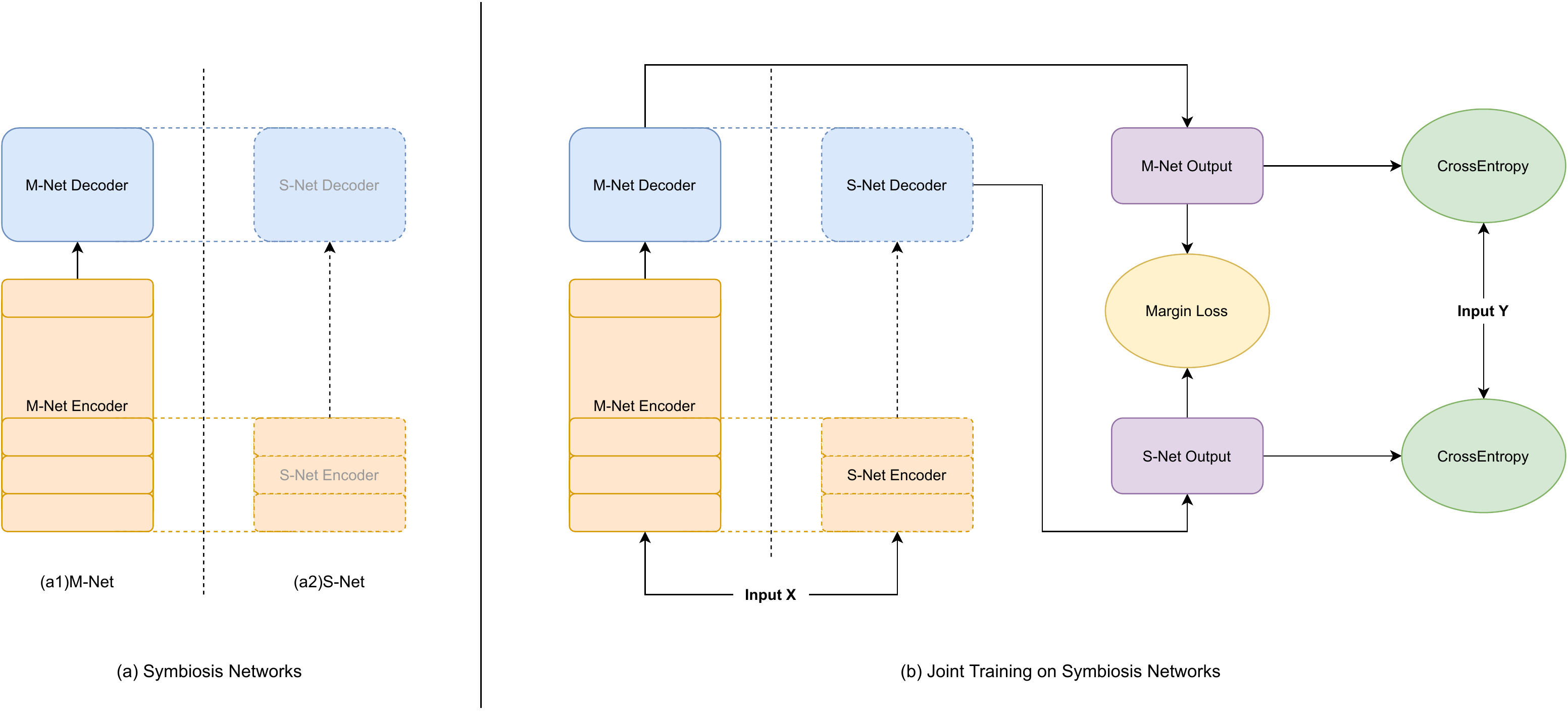} % Reduce the figure size so that it is slightly narrower than the column.
\caption{An overview of Symbiosis Networks and Joint-training on Symbiosis Networks.}
\label{fig:approach}
\end{figure*}

\subsection{Machine Translation}
\label{sec:bk_mt}

A machine translation task can be formally defined as a sequence to sequence generation problem: given the source sentence $X = \{x_1, x_2, ..., x_N\}$, to generate the target sentence $Y = \{y_1, y_2, ..., y_T\}$ according to the conditional probability $P(Y|X;\theta)$, where $\theta$ denotes the parameter set of a network. Different methods factorize the conditional probability differently.

The Transformer uses the auto-regressive factorization to minimize the following negative log-likelihood:

\begin{align}
\label{fun:bk_mt}
    \mathcal{L} = -log\ P(Y|X;\theta) = \sum_{t=1}^T -log\ p(y_t|y_{<t}, X; \theta)
\end{align}

, where $y_{<t} = \{[BOS], y_1, ..., y_{t-1}\}$.

\subsection{Transformer}

\paragraph{Vanilla Transformer}
Transformer-based model has achieved the state-of-the-art NMT system with the self-attention mechanism ~\citep{DBLP:conf/nips/VaswaniSPUJGKP17}. The Transformer architecture is a standard encoder-decoder model. The encoder side of Transformer can be viewed as a stack of $N$ layers containing a self-attention sub-layer and a feed-forward(FFN) sub-layer. The decoder shares a similar architecture as the encoder but possesses an encoder-decoder attention sub-layer to capture the mapping between two languages.

The standard attention used in the Transformer is the scaled dot-product attention. The input consists of queries and keys of dimension $d_k$, and values of dimension $d_v$. The dot products of the query with all keys are computed, scaled by $\sqrt{d_k}$, and a softmax function is applied to obtain the weights on the values. In practice, the attention function on a set of queries is computed simultaneously, packed together into a matrix $Q$. Assuming the keys and values are also packed together into matrices $K$ and $V$, the matrix of outputs is defined as:
\begin{align}
    Attention(Q,K,V) = softmax(\frac{QK^{T}}{\sqrt{d_k}})V
\end{align}

\paragraph{Deep Transformer}
Deep Transformer \citep{DBLP:conf/acl/WangLXZLWC19}, as the name implies, has the same architecture as Vanilla Transformer but stack more layers on the encoder side. For each sub-layer of Vanilla Transformer encoder, the computation pipeline is $\textit{self-attention/FFN} \rightarrow \textit{dropout} \rightarrow \textit{residual-add} \rightarrow \textit{normalize}$. This work flow places the layer normalization unit after self-attention/FNN and the residual connection. It is not efficient for training because layer normalization somehow blocks the information flow of the residual connection.

% TODO
A solution to the problem is to relocate layer normalization to remove the break between residual connections. Specifically, it is just to place the layer normalization unit before FNN/self-attention. The new computation
pipeline is $\textit{normalize} \rightarrow \textit{self-attention/FNN} \rightarrow \textit{dropout} \rightarrow \textit{residual-add}$. This model is defined as Pre-Norm Residual Network, and can establish a direct path from the bottom to the top of the network. Pre-norm residual network has been found to be more efficient for back-propagation over a large number of layers than the post-norm architecture. Deep Transformer is always a Pre-norm residual network.

\begin{algorithm*}[htp]
\begin{algorithmic}
\STATE 1: \textbf{procedure} JOINT\_TRAIN$(\mathcal{D} = \{X, Y\}, \mathcal{L})$: \\
\STATE 2: \hspace{\algorithmicindent} Train randomly initialized $(M^{M}, M^{S})$ on $\mathcal{D}$ optimized by $\mathcal{L}$ until convergence \\
\STATE 3: \hspace{\algorithmicindent} \textbf{Return} $(M^{M}, M^{S})$ \\
\STATE ~\\
\STATE 1: \textbf{procedure} MARGIN\_TRAIN$(M^{M}, M^{S}, \mathcal{D} = \{X, Y\}, \mathcal{L})$: \\
\STATE 2: \hspace{\algorithmicindent} Further train the pre-trained model $(M^{M}, M^{S})$ on $\mathcal{D}$ optimized by $\mathcal{L}$ until convergence \\
\STATE 3: \hspace{\algorithmicindent} \textbf{Return} $(M^{M}, M^{S})$ \\
\STATE~\\
\STATE 1: \textbf{procedure} SYM\_TRAIN($\mathcal{D} = \{X, Y\}$, $\mathcal{L}_{joint}$, $\mathcal{L}_{sym}$): \\
\STATE 2: \hspace{\algorithmicindent} $(M^{M}, M^{S}) \leftarrow $ JOINT\_TRAIN$(\mathcal{D}, \mathcal{L}_{joint})$ \\
\STATE 3: \hspace{\algorithmicindent} $(M^{M}, M^{S}) \leftarrow $ MARGIN\_TRAIN$(M^{M}, M^{S}, \mathcal{D}, \mathcal{L}_{sym})$\\
\STATE 4: \hspace{\algorithmicindent} \textbf{Return} $M^{M}$\\
\end{algorithmic}
\caption{Joint-training on Symbiosis Networks Algorithm}
\label{algo:train_algo}
\end{algorithm*}

\section{Approach}

The overall framework of our Joint-training on Symbiosis Networks for Deep Transformer is shown in Figure \ref{fig:approach}. We first present some necessary notations.

\subsection{Symbiosis Networks}
\label{sec:sn}

% Symbiotic Sub Network and Symbiotic Main Network
Symbiosis networks are a group of networks, including the whole network as the Symbiosis Main Network (M-Net) and its sub networks as Symbiotic Sub Networks (S-Nets). However, the S-Net is strictly defined. It is not pruned by randomly deleting some network nodes or modules from the M-Net. The S-Net have the same network structure as the M-Net, but partly different on the width or depth. The 
S-Net shares parameters with the M-Net. The learning objectives of the S-Net and the M-Net are the same. The purpose of Symbiosis networks is to enhance the main network.
In this work, we don't study multiple sub networks. We consider a M-Net and one S-Net in Symbiosis networks. In order to better describe the problem, we define shared part of M-Net as M-s-Net (s: shared), and the left part as M-i-Net (i: individual).

% Symbiosis Networks on MT
For neural machine translation, let $\mathcal{M}_{m-n}$ denotes the Deep Transformer model, where $m$ presents the number of encoder layers and $n$ presents the number of decoder layers. Then we can get Symbiosis networks on $\mathcal{M}_{m-n}$. The M-Net is $\mathcal{M}_{m-n}$, and the S-Net is $\mathcal{M}_{o-n}$ where $o<m$. 

$\mathcal{M}_{m-n}$ and $\mathcal{M}_{o-n}$ share parameters of word embedding layer, the whole $n$ decoder layers, and the first $o$ encoder layers. $\mathcal{M}_{m-n}$ and $\mathcal{M}_{o-n}$ have the same learning objective of minimize the negative log-likelihood \textbf{Eq} \ref{fun:bk_mt} in {\S}~\ref{sec:bk_mt}. When joint-training these two Symbiosis Transformer Networks, we aim to improve the performance of the M-Net $\mathcal{M}_{m-n}$.

\subsection{Joint-training on Symbiosis Networks}
\label{sec:train_sn}

Given the training data $\mathcal{D} = \{X, Y\}$, the basic learning objective for a deep machine translation model is to minimize the negative log-likelihood loss function, which is as follow:

\begin{align}
\label{fun:mt}
    \mathcal{L}_{mt} = -\log\mathcal{P}(Y|X;\theta)
\end{align}

Concretely, for Symbiotic networks, we can obtain two networks' parameters $\theta^{M}$ and $\theta^{S}$. Therefore, we can obtain two distributions of the model predictions, denoted as $\mathcal{P}(Y|X;\theta^{M})$ and $\mathcal{P}(Y|X;\theta^{S})$.

\begin{equation}
\begin{split}
\label{func_margin}
    \mathcal{L}_{\tau} &= \max\{0, \tau-\\
    &\bigg(\log \mathcal{P}(Y|X;\theta^{M}) - \log \mathcal{P}(Y|X;\theta^{S})\bigg)\}
\end{split}
\end{equation}

where $\log P(Y|X;\theta^{M}) - \log P(Y|X;\theta^{S})$ is the margin between the scores of the perfecter predicts and the imperfect predicts. The hinge loss on the margin encourages the log-likelihood of the predicts generated by Symbiotic Main Network to be at least $\tau$ larger than that of the imperfect predicts generated by Symbiotic Sub Network.

With the basic negative log-likelihood joint-learning objective $\mathcal{L}_{joint}$ of the two Symbiotic Networks:

\begin{align}
\label{func_mt_sym}
    \mathcal{L}_{joint} = -\frac{1}{2}\bigg(\log\mathcal{P}(Y|X;\theta^{M})
    +\log\mathcal{P}(Y|X;\theta^{S})\bigg)
\end{align}

the final training objective $\mathcal{L}_{sym}$ is to minimize for data $(X, Y)$:

\begin{equation}
\begin{split}
    \mathcal{L}_{sym} &= -\frac{1}{2}\bigg(\log\mathcal{P}(Y|X;\theta^{M}) +\log\mathcal{P}(Y|X;\theta^{S})\bigg) \\
    &+ \alpha \max\{0, \tau-\\
    &\bigg(\log \mathcal{P}(Y|X;\theta^{M}) - \log \mathcal{P}(Y|X;\theta^{S})\bigg)\}
\end{split}
\end{equation}

, where $\alpha$ is a hyper-parameter denoting the weight of $\mathcal{L}_{\tau}$ during the joint-training.

\subsection{Training Algorithm}

Let $M^{M}, M^{S}$ denotes the Symbiotic networks, where $M^{M}$ is the M-Net and $M^{S}$ is the S-Net. Let $\mathcal{D}$ presents the training dataset. Let $\mathcal{L}_{joint}$ and $\mathcal{L}_{sym}$ present the loss functions define in {\S}\ref{sec:train_sn}.

Our Joint-training on Symbiosis Networks strategy trains the models in two parts. In the first part, we train randomly initialized $(M^{M}, M^{S})$ on $\mathcal{D}$ optimized by $\mathcal{L}_{joint}$ until convergence. In the second part, we further train the $(M^{M}, M^{S})$ optimized by $\mathcal{L}_{sym}$, which adds an regularization loss $\mathcal{L}_{\tau}$.

For a clearer presentation, Algorithm \ref{algo:train_algo} summarizes the process concretely. We also call this strategy \textbf{Symbiosis Joint Training}.

\section{Experiments}

We evaluate our our proposed Symbiosis Joint Training strategy on three standard NMT benchmarks including WMT'14 En$\leftrightarrow$De in both directions, and WMT'14 En$\rightarrow$FR in single direction.

\begin{table*}[htp]
    \centering
    \resizebox{0.98\linewidth}{!}
    {
    \begin{tabular}{lcccccccc}
    \hline
    \textbf{Arch} & \textbf{M-Net} & \textbf{S-Net} & \multicolumn{6}{c}{\textbf{WMT'14}} \\
    \textbf{} & \textbf{} & \textbf{} & \multicolumn{2}{c}{\textbf{EN$\rightarrow$DE}} & \multicolumn{2}{c}{\textbf{DE$\rightarrow$EN}} & \multicolumn{2}{c}{\textbf{EN$\rightarrow$FR}} \\
     & & & \textbf{BLEU} & $\Delta$\textbf{} & \textbf{BLEU} & $\Delta$\textbf{} & \textbf{BLEU} & $\Delta$\textbf{} \\
    \hline
    \hline
    \multicolumn{7}{l}{\textbf{Baselines}} \\
    \hline
    \multicolumn{7}{l}{\textbf{Post-Norm}} \\
    Transformer-base\citep{DBLP:conf/nips/VaswaniSPUJGKP17} & 6-6 & - & 27.3 & - & - & - & - & - \\
    Transformer-big\citep{DBLP:conf/nips/VaswaniSPUJGKP17} & 6-6 & - & 28.4 & - & - & - & 41.8 & - \\
    \multicolumn{7}{l}{\textbf{Pre-Norm}} \\
    Transformer-base(Our Implementation) & 6-6 & - & 27.07 & - & 31.17 & - & 40.24 & - \\
    Transformer-big(Our Implementation) & 6-6 & - & 28.23 & - & 32.21 & - & 41.60 & - \\
    \hline
    \hline
    \multicolumn{7}{l}{\textbf{Classic Training}} \\
    \hline
    Transformer-deep & 12-6 & - & 27.57 & - & 31.78 & - & 40.83 & - \\
    Transformer-deep & 18-6 & - & 28.12 & - & 32.23 & - & 41.54 & - \\
    Transformer-deep & 21-6 & - & 28.25 & - & 32.26 & - & 41.73 & - \\
    Transformer-deep & 30-6 & - & 28.01 & - & 31.97 & - & 41.21 & - \\
    \hline
    \hline
    \multicolumn{7}{l}{\textbf{Symbiosis Joint Training}} \\
    \hline
    Transformer-deep & 12-6 & 6-6 & \textbf{28.18} & \textbf{+ 0.61} $\dagger$ & \textbf{32.27} & \textbf{+ 0.49} $\dagger$ & \textbf{41.52} & \textbf{+ 0.69} $\dagger$\\
    Transformer-deep & 18-6 & 6-6 & \textbf{28.49} & \textbf{+ 0.37} $\dagger$ & \textbf{32.46} & + 0.23 & \textbf{41.88} & + 0.34 \\
    Transformer-deep & 21-6 & 6-6 & \textbf{28.55} & + 0.30 & 32.45 & + 0.19 & 41.87 & + 0.14 \\
    Transformer-deep & 21-6 & 9-6 & 28.51 & + 0.26 & \textbf{32.47} & + 0.21 & \textbf{41.88} & + 0.15 \\
    Transformer-deep & 30-6 & 12-6 & \textbf{28.65} & + \textbf{0.64}$\dagger$ & \textbf{32.48} & + \textbf{0.51}$\dagger$ & \textbf{41.90} & + \textbf{0.69}$\dagger$ \\
    \hline
    \end{tabular}
    }
    \caption{Performance of BLEU \citep{DBLP:conf/acl/PapineniRWZ02} score under Symbiosis Joint Training strategy on WMT'14 En$\leftrightarrow$De and WMT'14 En$\rightarrow$FR.}
    \label{tab:results}
\end{table*}

\subsection{Settting}
\paragraph{Datasets}
The sizes of the datasets are (train=4.5M / valid=3k / test=3k / dict=42k) and (train=40.8M / valid=3k / test=3k / dict=45k) for WMT'14 En$\leftrightarrow$De and WMT'14 En$\rightarrow$FR respectively. We do the same pre-processing as previous work\citep{DBLP:conf/nips/VaswaniSPUJGKP17}, including \textit{clean-corpus-n}, \textit{normalize-punctuation} and \textit{remove-non-printing-char} with Moses toolkit\citep{DBLP:conf/acl/KoehnHBCFBCSMZDBCH07}. We tokenize each sentence into words by Moses, and further encode it into subwords by using BPE\citep{DBLP:conf/acl/SennrichHB16a}. 

\paragraph{Model Configurations}
We use Transformer\citep{DBLP:conf/nips/VaswaniSPUJGKP17} for all tasks. For Base/Deep model, the hidden size of attention layer is 512, and the size of FFN layer is 2,048. Also, we use 8 heads for attention. For training, we set all dropout to 0.1, including \textit{residual} dropout, \textit{attention} dropout, \textit{relu} dropout. Label smoothing $\epsilon_{ls} = 0.1$ is applied to enhance the generation ability of the model. For Big model, a larger hidden layer size (1,024), more attention heads (16), and a larger FFN layer dimensions (4,096) are required. The residual dropout was set to 0.3 for the EN$\leftrightarrow$DE tasks and 0.1 for the EN$\rightarrow$FR task.

\paragraph{Training}
Our models are trained on 8 Tesla V100 GPUs. Adam\citep{DBLP:journals/corr/KingmaB14} is used as the optimizer with $\beta_1 = 0.9$, $\beta_2 = 0.997$ and $\epsilon = 10^{-8}$. We adopt the same learning rate schedule. The learning rate (\textit{lr}) is first increased linearly for $\textit{warmup} =8,000$ steps from $1e^{-7}$ to $5e^{-4}$. After warmup, the learning rate decayed proportionally to the inverse square root of the current step. We batched sentence pairs by approximate length, and limited input/output tokens per batch to 4,096/GPU. All models are trained for 600k steps in total. For fair comparison, under our Symbiosis Joint Training, models are trained for 500k steps in the first stage and 100k steps in the second stage. The $\tau$ of $\mathcal{L}_{\tau}$ is set as 0.1 and the weighting parameter $\alpha$ is set as 1. We use fairseq\citep{ott2019fairseq} to do our experiments and employ FP16 to accelerate
training. And all results are the average of three times running with different random seeds.

\paragraph{Evaluation}
We use BLEU\citep{DBLP:conf/acl/PapineniRWZ02} as the evaluation metric. We average the last 6 consecutive checkpoints which are saved per training epoch on all WMT models. For all datasets, the length penalty is set to 0.6 and the beam size is set to 4.

\subsection{Results}

As shown in Table~\ref{tab:results}, our proposed Symbiosis Joint Training can stably improve the translation performance for Transformer-deep with multiple configurations comparing to the that of the same architecture under classic training.

For WMT'14 EN$\rightarrow$DE task, our Transformer-deep (12-6) model under Symbiosis Joint Training achieves \textbf{28.18} BLEU score, which significantly outperforms the same model under classic training by \textbf{0.61} BLEU and even outperforms the deeper model (18-6) under classic training. Our deepest model (21-6) under Symbiosis Joint Training forms the Transformer-base also achieves \textbf{0.30} BLEU improvement. 

For WMT'14 DE$\rightarrow$EN and EN$\rightarrow$FR task, we get similar conclusions. Specifically, our Transformer-deep (12-6) gains \textbf{0.49} BLEU improvement on DE$\rightarrow$EN task and  \textbf{0.69} BLEU improvement on EN$\rightarrow$FR. All results prove that, our proposed Symbiosis Joint Training strategy can improve the performance of the Symbiosis Main Network (M-Net).

\section{Study}

Beyond the superior experimental results, in this section, we conduct extensive studies on different perspectives to better understand our method. The analysis experiments are performed on the WMT14 En$\rightarrow$De translation task.

\subsection{Effect of the Regularization Loss $\mathcal{L}_{\tau}$}
\label{sec:study_reg}

We study the effect of the regularization loss $\mathcal{L}_{\tau}$. To make a fair comparison, we further train the models without loss $\mathcal{L}_{\tau}$ for 100k steps. The results show that, without loss $\mathcal{L}_{\tau}$ brings 0.15 BLEU drop on average. For detailed data, see Table \ref{tab:study_reg}.

\begin{table}[t]
\centering
\resizebox{0.98\linewidth}{!}
{
\begin{tabular}{@{}lccc@{}}
\hline
\textbf{Strategy} & \multicolumn{3}{c}{\textbf{Transformer-deep}} \\
\textbf{} & \textbf{12-6} & \textbf{18-6} & \textbf{21-6} \\
\hline
\hline
Baseline & {27.57} & {28.12} & {28.25} \\
\hline
w/ $\mathcal{L}_{\tau}$ & {28.18} & {28.49} & {28.55} \\
w/o $\mathcal{L}_{\tau}$ & \textbf{27.96} (-0.22) & \textbf{28.34} (-0.15) & \textbf{28.46} (-0.09) \\
\hline
\end{tabular}
}
\caption{BLEU scores of multiple models under w or w/o $\mathcal{L}_{\tau}$ training strategies on WMT14 EN$\rightarrow$DE task.}
\label{tab:study_reg}
\end{table}

\subsection{Effect of different S-Nets}
\label{sec:study_snets}

We design four strategies to build the Symbiotic Sub Networks (S-Nets) and study the effect of different S-Nets impact on M-Net. Let $\{\mathcal{H}_{0}^{M}, \mathcal{H}_{1}^{M}, ..., \mathcal{H}_{m-1}^{M}\}$ denotes $m$ M-Net encoder sub-layers, and $\mathcal{H}_{0}^{M}$ is the bottom sub-layer close to the embedding layer. Let $\{\mathcal{H}_{0}^{S}, \mathcal{H}_{1}^{S}, ..., \mathcal{H}_{n-1}^{S}\}$ denotes $n$ S-Net encoder sub-layers, where $n<m$. Also $\mathcal{H}_{0}^{S}$ is the bottom sub-layer. The strategies, including the baseline strategy are shown in as following.

\paragraph{Bottom Map:} The strategy of Bottom Map is to sequentially select the sub-layers starting from the bottom layer of M-Net to match the S-Net needs. \textbf{This is what we used in Symbiosis Joint Training}. $\mathcal{H}_{i}^{S} = \mathcal{H}_{i}^{M}$ where $i \in [0, n)$.

\paragraph{Top Map:} Contrary to Bottom Map, the strategy of Top Map is to sequentially select the sub-layers starting from the top layer of M-Net. $\mathcal{H}_{i}^{S} = \mathcal{H}_{m-n+i}^{M}$ where $i \in [0, n)$.

\paragraph{Top-Bottom Map:} Top-Bottom Map is the combination of Top Map and Bottom Map. For the top $n/2$ S-Net sub-layers order by Top Map, and the left sub-layers order by Bottom Map.

\paragraph{Linear Map:} Linear Map is, as the name implies, making a linear map between $m$ and $n$. $\mathcal{H}_{i}^{S} = \mathcal{H}_{i\times m \div n}^{M}$ where $i \in [0, n)$. Specifically, integer function may be applied to $i\times m \div n$ carefully.

\begin{table}[t]
\centering
\resizebox{0.98\linewidth}{!}
{
\begin{tabular}{@{}lcc@{}}
\hline
\textbf{Strategy} & \textbf{M-Net(12-6)} & \textbf{S-Net(6-6)} \\
\hline
\hline
\textbf{Baseline} & & \\
Base (6-6) & - & 27.07 \\
Deep (12-6) & 27.57 & - \\
\hline
\hline
Bottom Map & 28.18 & 26.94 \\
Top Map & 27.09 & 26.89 \\
Top-Bottom Map & 27.53 & 27.04 \\
Linear Map & 27.67 & 27.09 \\
\hline
\end{tabular}
}
\caption{BLEU scores of multiple models under w or w/o $\mathcal{L}_{\tau}$ training strategies on WMT14 EN$\rightarrow$DE task.}
\label{tab:study_snets}
\end{table}

We study the performance of these four strategies. We also evaluate the performance of the S-Nets. As shown in Table \ref{tab:study_snets}, Bottom Map achieves the best performance. Additionally, our Symbiosis Joint Training has little impact on the performance of S-Nets. All S-Nets achieve almost similar performance comparing with the Transformer-base baseline.

\section{Related Work}

\paragraph{LayerDrop}
LayerDrop proposed by \citep{DBLP:conf/iclr/FanGJ20}, applies structured dropout over layers during training, to make the model robust to pruned layers (shallower networks). Pruned layers can be regarded as sub layers. Although we all pay attention to sub-layers, our goals are quite different. They focus on model pruning and expect to achieve a competitive pruned network. But we focus on the main network and aim to improve the main network performance.

\paragraph{Shallow-to-Deep Training}
Shallow-to-Deep Training proposed by \citep{DBLP:conf/emnlp/LiWLJDXWZ20}, is a training method learns that deep models by stacking shallow models. In our Symbiosis Networks, the S-Net is similar to the shallow model. But we expect to study different questions. Their work aims to find an easy way to successfully train very deep models. But our purpose is to improve the deep model performance within limited stacks.

\section{Conclusion}

In this paper, we define Symbiosis Networks and present a new training strategy named Symbiosis Joint Training. The purpose of Symbiosis Networks is to enhance the main network. Furthermore, we adopt Symbiosis Networks on Transformer-deep model in NMT task. Apart from the basic learning objective for machine translation, another regularization loss can be designed to encourage the Symbiosis Main Network (M-Net) at least $\tau$ better than the Symbiosis Sub Network (S-Net). Our proposed training method improves Transformer-deep (12-6) by 0.60 BLEU over the baseline on average of multiple WMT14 translation tasks and even outperforms classic Transformer-deep (18-6).

\bibliography{acl2020,custom}

\begin{thebibliography}{14}
\expandafter\ifx\csname natexlab\endcsname\relax\def\natexlab#1{#1}\fi

\bibitem[{Bahdanau et~al.(2015)Bahdanau, Cho, and
  Bengio}]{DBLP:journals/corr/BahdanauCB14}
Dzmitry Bahdanau, Kyunghyun Cho, and Yoshua Bengio. 2015.
\newblock \href {http://arxiv.org/abs/1409.0473} {Neural machine translation by
  jointly learning to align and translate}.
\newblock In \emph{3rd International Conference on Learning Representations,
  {ICLR} 2015, San Diego, CA, USA, May 7-9, 2015, Conference Track
  Proceedings}.

\bibitem[{Fan et~al.(2020)Fan, Grave, and Joulin}]{DBLP:conf/iclr/FanGJ20}
Angela Fan, Edouard Grave, and Armand Joulin. 2020.
\newblock \href {https://openreview.net/forum?id=SylO2yStDr} {Reducing
  transformer depth on demand with structured dropout}.
\newblock In \emph{8th International Conference on Learning Representations,
  {ICLR} 2020, Addis Ababa, Ethiopia, April 26-30, 2020}. OpenReview.net.

\bibitem[{Kingma and Ba(2015)}]{DBLP:journals/corr/KingmaB14}
Diederik~P. Kingma and Jimmy Ba. 2015.
\newblock \href {http://arxiv.org/abs/1412.6980} {Adam: {A} method for
  stochastic optimization}.
\newblock In \emph{3rd International Conference on Learning Representations,
  {ICLR} 2015, San Diego, CA, USA, May 7-9, 2015, Conference Track
  Proceedings}.

\bibitem[{Koehn et~al.(2007)Koehn, Hoang, Birch, Callison{-}Burch, Federico,
  Bertoldi, Cowan, Shen, Moran, Zens, Dyer, Bojar, Constantin, and
  Herbst}]{DBLP:conf/acl/KoehnHBCFBCSMZDBCH07}
Philipp Koehn, Hieu Hoang, Alexandra Birch, Chris Callison{-}Burch, Marcello
  Federico, Nicola Bertoldi, Brooke Cowan, Wade Shen, Christine Moran, Richard
  Zens, Chris Dyer, Ondrej Bojar, Alexandra Constantin, and Evan Herbst. 2007.
\newblock \href {https://aclanthology.org/P07-2045/} {Moses: Open source
  toolkit for statistical machine translation}.
\newblock In \emph{{ACL} 2007, Proceedings of the 45th Annual Meeting of the
  Association for Computational Linguistics, June 23-30, 2007, Prague, Czech
  Republic}. The Association for Computational Linguistics.

\bibitem[{Li et~al.(2020)Li, Wang, Liu, Jiang, Du, Xiao, Wang, and
  Zhu}]{DBLP:conf/emnlp/LiWLJDXWZ20}
Bei Li, Ziyang Wang, Hui Liu, Yufan Jiang, Quan Du, Tong Xiao, Huizhen Wang,
  and Jingbo Zhu. 2020.
\newblock \href {https://doi.org/10.18653/v1/2020.emnlp-main.72}
  {Shallow-to-deep training for neural machine translation}.
\newblock In \emph{Proceedings of the 2020 Conference on Empirical Methods in
  Natural Language Processing, {EMNLP} 2020, Online, November 16-20, 2020},
  pages 995--1005. Association for Computational Linguistics.

\bibitem[{Ott et~al.(2019)Ott, Edunov, Baevski, Fan, Gross, Ng, Grangier, and
  Auli}]{ott2019fairseq}
Myle Ott, Sergey Edunov, Alexei Baevski, Angela Fan, Sam Gross, Nathan Ng,
  David Grangier, and Michael Auli. 2019.
\newblock fairseq: A fast, extensible toolkit for sequence modeling.
\newblock In \emph{Proceedings of NAACL-HLT 2019: Demonstrations}.

\bibitem[{Papineni et~al.(2002)Papineni, Roukos, Ward, and
  Zhu}]{DBLP:conf/acl/PapineniRWZ02}
Kishore Papineni, Salim Roukos, Todd Ward, and Wei{-}Jing Zhu. 2002.
\newblock \href {https://doi.org/10.3115/1073083.1073135} {Bleu: a method for
  automatic evaluation of machine translation}.
\newblock In \emph{Proceedings of the 40th Annual Meeting of the Association
  for Computational Linguistics, July 6-12, 2002, Philadelphia, PA, {USA}},
  pages 311--318. {ACL}.

\bibitem[{Sennrich et~al.(2016)Sennrich, Haddow, and
  Birch}]{DBLP:conf/acl/SennrichHB16a}
Rico Sennrich, Barry Haddow, and Alexandra Birch. 2016.
\newblock \href {https://doi.org/10.18653/v1/p16-1162} {Neural machine
  translation of rare words with subword units}.
\newblock In \emph{Proceedings of the 54th Annual Meeting of the Association
  for Computational Linguistics, {ACL} 2016, August 7-12, 2016, Berlin,
  Germany, Volume 1: Long Papers}. The Association for Computer Linguistics.

\bibitem[{Shaw et~al.(2018)Shaw, Uszkoreit, and
  Vaswani}]{DBLP:conf/naacl/ShawUV18}
Peter Shaw, Jakob Uszkoreit, and Ashish Vaswani. 2018.
\newblock \href {https://doi.org/10.18653/v1/n18-2074} {Self-attention with
  relative position representations}.
\newblock In \emph{Proceedings of the 2018 Conference of the North American
  Chapter of the Association for Computational Linguistics: Human Language
  Technologies, NAACL-HLT, New Orleans, Louisiana, USA, June 1-6, 2018, Volume
  2 (Short Papers)}, pages 464--468. Association for Computational Linguistics.

\bibitem[{Sutskever et~al.(2014)Sutskever, Vinyals, and
  Le}]{DBLP:conf/nips/SutskeverVL14}
Ilya Sutskever, Oriol Vinyals, and Quoc~V. Le. 2014.
\newblock \href
  {https://proceedings.neurips.cc/paper/2014/hash/a14ac55a4f27472c5d894ec1c3c743d2-Abstract.html}
  {Sequence to sequence learning with neural networks}.
\newblock In \emph{Advances in Neural Information Processing Systems 27: Annual
  Conference on Neural Information Processing Systems 2014, December 8-13 2014,
  Montreal, Quebec, Canada}, pages 3104--3112.

\bibitem[{Vaswani et~al.(2017)Vaswani, Shazeer, Parmar, Uszkoreit, Jones,
  Gomez, Kaiser, and Polosukhin}]{DBLP:conf/nips/VaswaniSPUJGKP17}
Ashish Vaswani, Noam Shazeer, Niki Parmar, Jakob Uszkoreit, Llion Jones,
  Aidan~N. Gomez, Lukasz Kaiser, and Illia Polosukhin. 2017.
\newblock \href
  {https://proceedings.neurips.cc/paper/2017/hash/3f5ee243547dee91fbd053c1c4a845aa-Abstract.html}
  {Attention is all you need}.
\newblock In \emph{Advances in Neural Information Processing Systems 30: Annual
  Conference on Neural Information Processing Systems 2017, December 4-9, 2017,
  Long Beach, CA, {USA}}, pages 5998--6008.

\bibitem[{Wang et~al.(2019)Wang, Li, Xiao, Zhu, Li, Wong, and
  Chao}]{DBLP:conf/acl/WangLXZLWC19}
Qiang Wang, Bei Li, Tong Xiao, Jingbo Zhu, Changliang Li, Derek~F. Wong, and
  Lidia~S. Chao. 2019.
\newblock \href {https://doi.org/10.18653/v1/p19-1176} {Learning deep
  transformer models for machine translation}.
\newblock In \emph{Proceedings of the 57th Conference of the Association for
  Computational Linguistics, {ACL} 2019, Florence, Italy, July 28- August 2,
  2019, Volume 1: Long Papers}, pages 1810--1822. Association for Computational
  Linguistics.

\bibitem[{Wu et~al.(2016)Wu, Schuster, Chen, Le, Norouzi, Macherey, Krikun,
  Cao, Gao, Macherey, Klingner, Shah, Johnson, Liu, Kaiser, Gouws, Kato, Kudo,
  Kazawa, Stevens, Kurian, Patil, Wang, Young, Smith, Riesa, Rudnick, Vinyals,
  Corrado, Hughes, and Dean}]{DBLP:journals/corr/WuSCLNMKCGMKSJL16}
Yonghui Wu, Mike Schuster, Zhifeng Chen, Quoc~V. Le, Mohammad Norouzi, Wolfgang
  Macherey, Maxim Krikun, Yuan Cao, Qin Gao, Klaus Macherey, Jeff Klingner,
  Apurva Shah, Melvin Johnson, Xiaobing Liu, Lukasz Kaiser, Stephan Gouws,
  Yoshikiyo Kato, Taku Kudo, Hideto Kazawa, Keith Stevens, George Kurian,
  Nishant Patil, Wei Wang, Cliff Young, Jason Smith, Jason Riesa, Alex Rudnick,
  Oriol Vinyals, Greg Corrado, Macduff Hughes, and Jeffrey Dean. 2016.
\newblock \href {http://arxiv.org/abs/1609.08144} {Google's neural machine
  translation system: Bridging the gap between human and machine translation}.
\newblock \emph{CoRR}, abs/1609.08144.

\bibitem[{Yang et~al.(2021)Yang, Ma, Zhang, Wan, Li, and
  Zhou}]{DBLP:conf/naacl/YangMZWLZ21}
Jian Yang, Shuming Ma, Dongdong Zhang, Juncheng Wan, Zhoujun Li, and Ming Zhou.
  2021.
\newblock \href {https://doi.org/10.18653/v1/2021.naacl-main.312} {Smart-start
  decoding for neural machine translation}.
\newblock In \emph{Proceedings of the 2021 Conference of the North American
  Chapter of the Association for Computational Linguistics: Human Language
  Technologies, {NAACL-HLT} 2021, Online, June 6-11, 2021}, pages 3982--3988.
  Association for Computational Linguistics.

\end{thebibliography}
\bibliographystyle{acl_natbib}

\end{document}